\documentclass[11pt,a4paper]{article}
\usepackage{authblk}
\usepackage[hyperref]{acl2020}
\usepackage{times}
\usepackage{latexsym}

\usepackage{url}
\usepackage{booktabs}
\usepackage{graphicx}
\usepackage{amsmath}
\usepackage{mathtools, cuted}
\usepackage{inconsolata} 
\usepackage{makecell}
\usepackage{multirow}

\usepackage{microtype}

\aclfinalcopy 

\title{Local word statistics affect reading times independently of surprisal}

\author[$\ast$]{\textbf{Adam Goodkind}}
\author[$\ddagger,\ast$]{\textbf{Klinton Bicknell}}
\affil[$\ast$]{Northwestern University \quad $^\ddagger$Duolingo}
\affil[ ]{\texttt{a.goodkind@u.northwestern.edu}}
\affil[ ]{\texttt{klinton@duolingo.com}}

\date{}

\begin{document}
\maketitle

\begin{abstract}
Surprisal theory has provided a unifying framework for understanding many phenomena in sentence processing \citep{hale2001probabilistic,levy2008expectation}, positing that a word's conditional probability given all prior context fully determines processing difficulty. Problematically for this claim, one local statistic, word frequency, has also been shown to affect processing, even when conditional probability given context is held constant. Here, we ask whether other local statistics have a role in processing, or whether word frequency is a special case. We present the first clear evidence that more complex local statistics, word bigram and trigram probability, also affect processing independently of surprisal. These findings suggest a significant and independent role of local statistics in processing. Further, it motivates research into new generalizations of surprisal that can also explain why local statistical information should have an outsized effect.
\end{abstract}

\section{Introduction}
\label{sec:introductiion}

In the field of human sentence processing, surprisal theory \citep{hale2001probabilistic,levy2008expectation} offers a successful unifying framework capturing a wide range of phenomena, including syntactically locally ambiguous ``garden path'' sentences, pronoun ambiguity resolution, and word predictability effects. The theory is centered around the intuitive notion that words are more difficult to process when they are less likely to occur in their context. More formally, surprisal theory proposes that difficulty in processing word $w_n$ is proportional to its negative log probability conditional on all previous words:

\begin{equation} \label{eq:surprisal}
\mbox{Difficulty}(w_n) \propto -\log p(w_n|w_1, \dots, w_{n-1})
\end{equation}

In making this statement, surprisal theory implies that the probability of a word given preceding context is a \emph{causal bottleneck} in linking linguistic properties to processing difficulty \citep[Sec.~2.3]{levy2008expectation}. That is, other linguistic properties studied in the sentence processing literature, e.g., syntax or pragmatics, affect processing difficulty by affecting conditional probability, which then affects surprisal. In other words, all linguistic properties ultimately flow through and contribute to surprisal, and surprisal determines final processing difficulty.
This linking hypothesis with reading times can be derived as a prediction of an optimal, unconstrained comprehension system \citep{levy2008expectation,smith2013predictability}. Thus, to the extent that surprisal captures all effects of linguistic context on reading times, it is a signature of a fully \emph{rational} sentence processor that makes full use of linguistic context.

Despite the success of surprisal theory, there is, however, reason to suspect that this causal bottleneck may not empirically hold. Instead, comprehenders may exhibit special sensitivity to local information, even when the global probability given context (surprisal) is held constant. In the syntactic and semantic domains, this has been studied under the name of ``local coherence'' effects \citep[e.g.,][]{tabor2004effects,konieczny2005psychological}. Results there have suggested that comprehenders may sometimes interpret short phrases embedded in a sentence in ways that are inconsistent with that sentence context, but consistent with how that phrase is frequently used.

Here, we investigate a related question of whether local word transition statistics, such as those captured by $n$-gram models, affect processing independently of surprisal. Relevant to this question, it is already clear that one local statistic, word frequency or the marginal $p(w_n)$, does influence reading times independently of surprisal: if two words are equally likely in context, the more frequent word will still be read faster \citep[see e.g.][among many others, but see \citealp{shain2019large} for a contrasting result]{broadbent1967word,rayner1986lexical}. Such results may suggest that language comprehenders may have special sensitivity to  local statistics broadly, and would also exhibit faster reading times for frequent bigrams or trigrams even when surprisal given the full context  was matched. However, it is also possible (and, we suspect, is often assumed) that word frequency has an effect not because it is a local statistic, but for some other reason idiosyncratic to word frequency. For example, perhaps this effect is related to the cognitive mechanisms of individual word retrieval, or even word learning.

In this paper, we tease apart these two possibilities, and present what we believe is the first clear evidence for effects on reading times of more complex local word statistics, bigram $p(w_n|w_{n-1})$ and trigram $p(w_n|w_{n-1},w_{n-2})$ probability, over and above a baseline of word frequency plus surprisal given the full context. To do this, we first build a state-of-the-art language model to compute the best possible estimates of surprisal given prior context. We then use surprisal computed from this language model to predict reading times in a corpus of eye movements in reading via mixed-effects regression, and show that word bigram and trigram (log) probability predictors have significant effects above and beyond the surprisal baseline, after controlling for known covariates.\footnote{All language models and regression models code is available at \url{https://github.com/langcomp/lm_1b}}

Our paper is structured as follows: 
In Section \ref{sec:prior} we summarize prior work on the effects of word \textit{n}-grams in human sentence processing, and we describe a concern that prevents these studies from providing strong evidence for \textit{n}-gram effects above and beyond a surprisal baseline. In Section \ref{sec:experiment1} we describe the state-of-the-art language model that we use to estimate word surprisal, and demonstrate that the concern raised about prior work only mildly applies to the present surprisal model. Section \ref{sec:experiment2} shows that bigram and trigram probabilities significantly affects sentence processing above and beyond surprisal. Finally, we discuss these results and look to future directions.

\section{Local word statistics in processing}
\label{sec:prior}

A number of studies have investigated the processing effects of local statistics such as word frequency and bigram probability, but only in isolation. For example, \citet{mcdonald2003low} studied a corpus of eye movements and found a significant effect of bigram probabilities  (aka ``transitional probabilities'' in these studies ) on both first fixation and gaze duration. \citet{mcdonald2003eye} then performed two eye-tracking studies that similarly found transitional probability and word frequency to have a significant effect on eye movement.

\citet{arnon2010more} looked at four-gram multi-word expressions (MWEs) and found that comprehenders' reaction times were sensitive to the frequencies of compositional four-word phrases. Importantly for their finding, the effect of the four-word phrase frequencies could not be reduced to the frequency of individual words. 

Although these studies all documented effects of local statistics, they did not control for surprisal. Because of this, a reasonable explanation for the results observed is that the bigram and four-gram effects were actually (somewhat crude) estimates of surprisal given the full context, since the full context does include the words immediately prior. 

It is also the case, however, that some studies have reported effects of bigram probability above and beyond surprisal \citep{demberg2008data,fossum2012sequential,mitchell2010syntactic} -- although this point was not a focus of these studies. However, we argue that these results may actually not be a problem for surprisal theory because of the way surprisal was estimated. In each of these studies, the language model used to estimate surprisal was a probabilistic context-free grammar (PCFG). PCFGs are syntactic language models that include an independence assumption that effectively means that relationships between particular words cannot be learned, except for those mediated by parts-of-speech. For example, a PCFG could learn that a verb is likely to follow a noun in a particular context, but not that the verb `meowed' is likely to follow the noun `cat'. 
    
This restriction means that the PCFG language model used in the prior work to estimate surprisal conditioning on the full context was effectively only conditioning on the full context of parts-of-speech. By contrast, although the bigram probability was only conditioning on one word of (local) context, it could use that word's identity to make predictions instead of just its part-of-speech. Thus, instead of these bigram effects being clear evidence for local context having an outsized effect above and beyond global context, they may be more likely merely evidence that the \emph{identities} of prior words are important in language modeling.

In this vein, \citet{frank2009surprisal} and \citet{monsalve2012lexical} show that language models built with neural networks and \textit{n}-grams perform better than PCFGs. In light of the argument above, this is unsurprising, since a model built on a PCFG will not predict word identities as well as a model built from neural networks or \textit{n}-grams.

A more formal statement of this point is that estimation error across language models is somewhat independent, especially if the two language models are different classes of model, as was the case for PCFGs and bigrams. In this situation, taking a weighted average (or interpolation) of the predictions of two language models can often yield a better language model than either individually. An interpolation $p_{interp}$ of two language models $p_1$ and $p_2$ is formally defined as a linear combination with interpolation parameter $\gamma$:

\begin{equation} \label{eq:interp}
\begin{split}
p_{interp}(w_n|w_1^{n-1}) = \\ \gamma p_1(w_n|w_1^{n-1}) 
+ (1-\gamma) p_2(w_n|w_1^{n-1})
\end{split}
\end{equation}

Similarly, simultaneous effects of PCFG surprisal and bigram probability in a regression model may actually result from the regression model implicitly interpolating the PCFG and bigram language model probabilities to form a better estimate of a word's actual probability in context (surprisal), where the regression coefficients correspond to an unnormalized version of $\gamma$.

In the present work, we provide a stronger test of whether word bigram and trigram probabilities have effects on reading times above and beyond surprisal. We do this by estimating surprisal with a state-of-the-art word-based language model, which does condition on prior word identities, unlike a PCFG. 
We then verify in Experiment 1 that this language model can only be improved by a limited amount by interpolating surprisal probabilities with \textit{n}-gram probabilities. We then show in Experiment 2 that reading times are dependent upon bigram and trigram probabilities to a much greater extent than than can be explained by optimal interpolation.

\section{Experiment 1: Perplexity analysis}
\label{sec:experiment1}

Our first experiment sought to measure the extent to which interpolating a state-of-the-art word-based language model with word \textit{n}-gram models would improve the language model's predictions for language (i.e., its perplexity). The measurements can then be compared to those in Experiment 2, to test whether there are outsized effects of local context (bigrams and trigrams) in human sentence processing.

\subsection{Methods}
\label{ssec:exp1methods}

All language models were estimated from the Google One Billion Word Benchmark corpus \cite{chelba2013billion}. This corpus is made up of about 800 million word tokens collected from international English newswires, with a unique vocabulary size of approximately 800,000 words. 

\subsubsection{Language model construction}

The language model we used to estimate surprisal follows \citet{goodkind2018predictive}, who built a model with state-of-the-art surprisal estimates by interpolating the off-the-shelf Long Short-term Memory (LSTM)-based language model described and estimated by \citet{jozefowicz2016exploring} with a 5-gram model. We used the same interpolation weight as \citet{goodkind2018predictive}, since they reported it minimized perplexity on the Dundee corpus, which we use in Expt.~2. 

We estimated all $n$-gram models with interpolated modified Kneser-Ney smoothing and default parameters using \texttt{kenlm} \cite{heafield2013scalable}.

\begin{figure}
	\includegraphics[width=.45\textwidth]{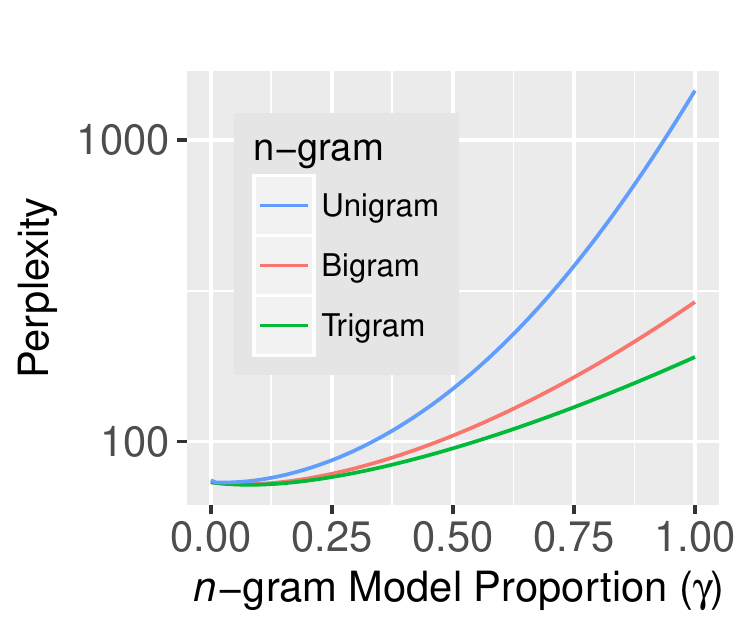}
    \caption{Perplexity for interpolations of the surprisal model with unigram, bigram, and trigram models. The y-axis uses a log-scale. The best (lowest) perplexity comes from using a small proportion of the \textit{n}-gram language models. The exact proportions are described in Section \ref{ssec:exp1results}. The full \textit{n}-gram models, at the rightmost edge of the graph where $\gamma=1.00$, provides the least accurate perplexity.}
    \label{fig:interp}
\end{figure}

\subsubsection{Interpolation experiment}

The specific goal of this experiment, then, was to determine the extent to which interpolating the language model we used to estimate surprisal with $n$-gram models would improve its perplexities on the Dundee corpus. Specifically, we interpolated it with a unigram (frequency-based) model, a bigram model, and a trigram model. In each, we tested interpolation parameter values ($\gamma$ in Figure \ref{fig:interp}) from 0 to 1 in increments of 0.01. 

The most commonly used form of probabilistic interpolation, which we will call \textit{additive} interpolation, combines language model probabilities by means of adding together complementary proportions of probabilities from each model (see Eq.~\ref{eq:interp}). However, prior work has established that log probabilities (surprisal) have a more linear relationship with reading times than raw probabilities \cite{smith2013predictability}. For this reason, the regression we perform in Experiment 2 uses log probabilities as predictors, which means that the effective interpolation that happens in such a regression is an interpolation of \emph{log} probabilities, rather than raw probabilities. This is equivalent to a type of weighted multiplication of raw probabilities, so we here call it \textit{multiplicative} interpolation. 

Because multiplying probabilities together does not keep the total probability mass constant as in additive interpolation, multiplicative interpolation must be normalized as an additional step. The first step of multiplicative interpolation (Eq.~\ref{eq:mult_interp}), which yields an unnormalized log probability $\tilde{p}$, can be written parallel to Eq.~\ref{eq:interp}. This then must be normalized to yield the actual interpolated log probability (Eq.~\ref{eq:logsumexp}), where the sum is taken over all vocabulary items. This normalization step must be done separately for each context.

\begin{equation} \label{eq:mult_interp}
\begin{split}
\log \tilde{p}_{interp}(w_n|w_1^{n-1}) = \gamma \log p_1(w_n|w_1^{n-1}) + \\ (1-\gamma) \log p_2(w_n|w_1^{n-1})
\end{split}
\end{equation}

\begin{equation} \label{eq:logsumexp}
\begin{split}
\log p_{interp}(w_n|w_1^{n-1})
 = \log \tilde{p}_{interp}(w_n|w_1^{n-1}) -\\ \log \sum_{w \in V} \tilde{p}_{interp}(w|w_1^{n-1})
\end{split}
\end{equation}

\subsection{Results}
\label{ssec:exp1results}

Figure \ref{fig:interp} illustrates the results of multiplicative interpolation. For each interpolated mixture, a small amount of \textit{n}-gram weighting improves overall perplexity. This weighting is only $0.02$ for unigrams, $0.07$ for bigrams, and $0.08$ for trigrams. 

\begin{table}
\centering
\begin{tabular}{lccr}
\toprule
Model & \thead{Optimal \textit{n}-gram\\Proportion} & \thead{Optimal\\Perplexity} \\ \midrule
Unigram & 0.02 & 73 \\
Bigram & 0.07 & 72 \\
Trigram & 0.08 & 72 \\ \bottomrule
\end{tabular}%
\caption{Results of Experiment 1, represented graphically in Figure \ref{fig:interp}. The best perplexity was reached using the optimal surprisal proportion. However, the full surprisal model achieved a perplexity of $\sim$73.3, which is very close to the optimal levels reported.}
\label{tbl:exp1}
\end{table}

From this, we can conclude that adding in a small amount of unigram, bigram or trigram probability information improves the accuracy of our state-of-the-art surprisal language model, although not by much.\footnote{It still remains undetermined \textit{why} this small amount improves accuracy. However, in an unreported experiment using standard (additive) interpolation, the \textit{n}-gram models do not improve perplexity at all, and the full proportion of the surprisal model achieves the lowest perplexity.}
From this, we can proceed to Expt.~2, where any effects we see from local statistics larger than these effects reported in Expt.~1 cannot be explained by a deficiency in language model surprisal estimates.

\section{Experiment 2: Gaze duration analysis}
\label{sec:experiment2}

While Experiment 1 found that a small amount of information from trigram, bigram and unigram probabilities improves the perplexity of our language model's predictions of language, we wanted to examine whether these local statistics had an even greater effect on reading times (gaze duration) over and above a surprisal baseline.

\subsection{Methods}


\subsubsection{Dataset}
We analyzed reading times taken from the Dundee Eye-Tracking Corpus \citep{kennedy2003dundee}, which recorded the eye-movement data from 10 English-speaking participants reading newspaper editorials in \textit{The Independent}, a British newspaper.

For the present study specifically, we predict gaze durations for each word, defined to be the sum of all fixations made on a word between the time the word is initially fixed and when the eyes first move off of the word. This measure is only calculated if the word is fixated by that reader prior to any fixation on a later word (i.e., during `first pass' reading).\footnote{We limit our study to the first pass because the surprisal theory is only about the influence of the prior context on processing. 
If the word was not fixated during first pass reading, this is missing data. We used a total of about 436,000 valid gaze durations in the English portion of the Dundee corpus. After performing the exclusions listed below, we were left with a total of 289,726 gaze duration measurements and a vocabulary size of 37,420 word types.}

In line with previous studies of gaze durations using the Dundee Corpus \cite[e.g., ][]{demberg2008data,smith2013predictability,goodkind2018predictive}, we excluded words preceding punctuation, words with non-alphabetical characters, and words that were presented to participants at the beginning or end of a line of text. We also omitted any words that were outside the vocabulary (OOV) of the 1b corpus (and thus the language model).


Further, because our statistical model of the gaze duration of each word also included effects of the preceding word, we also excluded words where the preceding word was not informative. These included words following punctuation, and words that \textit{followed} words with non-alphabetical characters and OOV words.

\subsubsection{Statistical models}

To test the strength of our predictors of interest we constructed generalized linear mixed-effect models (GLMMs) with the \texttt{lme4} package in \texttt{R}, reporting the coefficients of the predictors \cite{bates2015lme4,r2013r}. The predictors of interest for our regression models were the unigram, bigram, and trigram log probability, as well as surprisal. Both the current and previous word, $w_{n}$ and $w_{n-1}$, were included. These were then centered on the mean. 

Our first model only used a unigram predictor, and this model then served as a baseline. The bigram model added a bigram predictor, and the trigram model added only a trigram predictor to the unigram baseline. Because of the high degree of correlation between bigram and trigram probability (e.g., in our dataset Spearman's $\rho = 0.92$), we test their effects separately. 

Similar to \citet{smith2013predictability}, we also used generalized additive mixed-effects models (GAMMs) to predict reading times with the \texttt{mgcv} \cite{wood2004mgcv} package in \texttt{R}. GAMMs allow us to include arbitrary non-linear effects of other covariates. By including non-linear effects, we provide corroborating support for the coefficients reported in the GLMMs, in that the GAMM coefficients demonstrate the GLMMs are not interpolating just to account for non-linear surprisal effects.

To control for other known effects in eye-tracking that were not of interest in the present study, we added a number of covariates to our regression models. The covariates were identical to those in \citet{goodkind2018predictive}, and very similar to \citet{smith2013predictability}. For both the current and previous word, these included tensor product interactions for (log) orthographic word length and word frequency, as well as random slopes for \textit{n}-gram probability by subject. We also included a spline effect that reflected the word number within the text ($\nu_n$). Finally, we included a binary variable of whether or not the previous word had received a fixation ($\pi_n$).
The GLMMs used the same predictors as the GAMMs, except that the GLMM covariates were all linear. 


The full model formulas used are given in Figures \ref{fig:glm-formula} (GLMM) and \ref{fig:gam-formula} (GAMM), in \texttt{lme4} and \texttt{mgcv} syntax respectively. The formulas presented in Figure \ref{fig:glm-formula} and Figure \ref{fig:gam-formula} are for a bigram model. For the unigram baseline, the bigram terms would be removed. For the trigram models, the bigram terms would be replaced with trigrams.

\begin{figure*}[ht]
    \begin{math}
    \begin{aligned}
    \small{\texttt{gaze duration $\sim$ surprisal$_n$ + surprisal$_{n-1}$ + freq$_n$ + freq$_{n-1}$ + bigram$_n$ + bigram$_{n-1}$ +}}\\[-2mm]
    \small{\texttt{freq$_n$:length$_n$ + freq$_{n-1}$:length$_{n-1}$ + (freq$_n$ + freq$_{n-1}$ +}}\\[-2mm]
    \small{\texttt{bigram$_n$ + bigram$_{n-1}${||}subject) + 
    $\pi_n$ + $\nu_n$}}
    \end{aligned}
    \end{math}
    \caption{Formula for generalized linear mixed-effect bigram model in \texttt{lme4} syntax}
    \label{fig:glm-formula}
\end{figure*}

\begin{figure*}[ht]
    \begin{math}
    \begin{aligned}
    \small{\texttt{gaze duration $\sim{}$ s(surprisal$_n$, bs='cr', k=40) + 
    s(surprisal$_{n-1}$, bs='cr', k=40) + }}\\[-2mm]
    \small{\texttt{te(freq$_n$, log(length$_n$), bs='tp') +
    te(freq$_{n-1}$, log(length$_{n-1}$), bs='tp') +
    $\pi_n$ + s($\nu_n$, bs='cr') + }}\\[-2mm]
    \small{\texttt{bigram$_n$ + bigram$_{n-1}$ + 
    s(bigram$_n$, subject, bs='re') +
    s(bigram$_{n-1}$, subject, bs='re') + }}\\[-2mm]
    \small{\texttt{s(freq$_n$, subject, bs='re') +
    s(freq$_{n-1}$, subject, bs='re') +
    s(subject, bs='re')}}
    \end{aligned}
    \end{math}
    \caption{Formula for generalized additive mixed-effect bigram model in \texttt{mgcv} syntax}
    \label{fig:gam-formula}
\end{figure*}

\subsection{Results}

We assessed the significance of predictors of interest via likelihood ratio tests (LRT) for the GLMMs and using $p$-values derived from the $t$ statistics for the GAMMs. The estimated coefficient values and $p$-values can be seen in Table \ref{tbl:exp2_coef}.

\begin{table*}
\centering
\begin{tabular}{@{}lrrrrr@{}}
\toprule
Predictor & \multicolumn{2}{c}{GLMM $\hat{\beta}$} & \multicolumn{2}{c}{GAMM $\hat{\beta}$} & Surprisal $\hat{\beta}$ \\ \midrule
log frequency $w_n$ & -12.40 & $p<0.01$ & -10.42 & $p<0.01$ & -3.05 \\
log frequency $w_{n-1}$ & -2.41 & $p<0.01$ & -2.83 & $p<0.001$ & -4.34 \\
log bigram probability $w_n$ & -1.49 & $p<0.05$ & -1.13 & $p=0.09$ & -2.30 \\
log bigram probability $w_{n-1}$ & -0.74 & $p=0.08$ & -1.09 & $p<0.05$ & -4.03 \\
log trigram probability $w_n$ & -1.52 & $p<0.05$ & -1.38 & $p<0.05$ & -1.94 \\
log trigram probability $w_{n-1}$ & -1.22 & $p<0.01$ & -1.14 & $p<0.05$ & -3.50 \\ \bottomrule
\end{tabular}%
\caption{Coefficients for all predictors of interest, along with \textit{p}-values for the significance of the predictor. For each model, there are two rows of results: the first row provides the results for the current word; the second row provides the results for the previous word.}
\label{tbl:exp2_coef}
\end{table*}

The unigram (frequency) predictors of both the current and previous word had significant effects on gaze duration for both GLMM and GAMM analyses. This is unsurprising given previous findings that frequency affects reading times over and above surprisal.

More notably, we also found significant effects of the bigram and trigram predictors. Specifically, for the GLMM analysis, there was a significant effect of current bigram probability ($p\mbox{s}<0.05$), and marginal significance for the previous bigram ($p\mbox{s} = 0.08$). In the stronger baseline of the GAMM, the current bigram was marginally significant ($p\mbox{s} = 0.09$), while the previous bigram was significant ($p\mbox{s}<0.05$). 

Finally, all trigram predictors $p$-values were significant for both the GLMM and GAMM. This provides a much more clear picture for the effects of trigrams on reading times over and above surprisal and a word frequency baseline.

\subsubsection{Inferred Interpolation Weights}
\label{sssec:inferred_interp}

Now that we have established that bigrams and trigrams have significant effects, we must ask whether these effects are larger than the results of Experiment 1 could explain as simply optimal multiplicative interpolation. Specifically, we use the coefficients of the GLMMs to infer effective interpolation weights from each regression model. We measured this as a proportion of coefficients (Equation \ref{eq:infer_interp}), where the \textit{n}-gram coefficient $\beta_n$ is divided by the sum of the surprisal coefficient $\beta_s$ and the \textit{n}-gram coefficient. 
We can then compare these effective interpolation weights and perplexities to the optimal weights and perplexities in Experiment 1, to understand whether the regression models measuring human language processing utilize a greater proportion of \textit{n}-gram information than that which achieved optimal perplexity on textual data.

\begin{equation} \label{eq:infer_interp}
\frac{\beta_n}{\beta_s + \beta_n}
\end{equation}

\begin{table*}
\centering
\begin{tabular}{@{}lrrrr@{}}
\toprule
Predictor & \thead{Effective\\Interpolation} & \thead{Optimal\\Interpolation} & \thead{Effective\\Perplexity} & \thead{Optimal\\Perplexity} \\ \midrule
log frequency $w_n$ & 0.80 & 0.02 & 484 & 73 \\
log frequency $w_{n-1}$ & 0.36 & 0.02 & 105 & 73 \\
log bigram probability $w_n$ & 0.39 & 0.07 & 90 & 72 \\
log bigram probability $w_{n-1}$ & 0.16 & 0.07 & 74 & 72 \\
log trigram probability $w_n$ & 0.44 & 0.08 & 89 & 72 \\
log trigram probability $w_{n-1}$ & 0.26 & 0.08 & 77 & 72 \\ \bottomrule
\end{tabular}%
\caption{The effective interpolation uses the formula in Eq.~\ref{eq:infer_interp} to calculate interpolation weighting for gaze duration. The optimal interpolation and corresponding perplexities were calculated in Expt.~1 and reported in Table \ref{tbl:exp1}, reproduced here in order to facilitate comparison between effective and optimal interpolation weights. We also provide the perplexity at the effective weighting level, for comparison to the optimal perplexity achieved in Expt.~1.}
\label{tbl:exp2_interp}
\end{table*}

As can be seen in Table \ref{tbl:exp2_interp}, the regression models based on gaze duration all utilize a greater proportion of \textit{n}-gram weighting than the optimal interpolation results of Experiment 1. However, in some cases, the effective interpolation is not much higher than optimal, and would yield a perplexity that is near-optimal, so we go through each of the 6 coefficients separately.

For unigrams, effective interpolation weights for both the current and prior word are far higher than optimal, 0.8 and 0.36 effective versus 0.02 optimal. Interpolating so heavily with a unigram model unsurprisingly yields very high perplexities of 484 and 105 versus 73 for the optimal model. These results replicate prior work in showing robust effects of unigram probability (word frequency) that cannot be explained as merely optimal interpolation.

For bigrams and trigrams, the story is more complex, and intriguingly depends whether effects of the current or prior word are being examined. For the current word, the regression model's effective interpolation weights for bigram and trigram probability are again substantially higher than optimal interpolation, at 0.39 and 0.44 respectively, versus 0.07 and 0.08 for optimal interpolation. Weighting bigram and trigram probability models yields perplexities that are about 25\% worse than optimal perplexity -- at 90 and 89 versus 72 for both optimal interpolations -- and so cannot be explained as simply a version of optimal interpolation. These results provide strong evidence that the current word's bigram and/or trigram probability has significant effects on gaze durations above and beyond surprisal.

For effects of the prior word's bigram and trigram probability, however, the situation appears to be different. Here, the regression models' effective interpolation weights for bigram and trigram probability are just 0.16 and 0.26, much closer to the optimal weightings of 0.07 and 0.08. 

While these interpolation weights are more than 2 times the optimal weighting (which achieves the best perplexity in Expt. 1), they only result in perplexities here that are about 2--7\% higher, 74 and 77 versus 72 for both optimal models. These results suggest that the effective interpolation of bigram/trigram information with surprisal for effects of the prior word may be explainable as simply optimal interpolation, or at least very close to it. The apparent difference between the situation here with prior word effects and that for current word effects may suggest something about the timecourse of effects of local statistics in processing. Perhaps local information has a role early in processing, but lingering effects of a word are only related to its actual surprisal. 

\section{Discussion}
\label{sec:discussion}

Combining the results of Experiments 1 and 2 provides strong evidence for effects of local statistics on sentence processing. Our first experiment measured the extent to which the language model we used to estimate surprisal could be improved by overweighting local statistics. We measured this by multiplicatively interpolating this language model with three different $n$-gram models and found that a small proportion of an $n$-gram model produces a language model with slightly more accurate predictions, as measured by perplexity. This was in contrast to prior work that used PCFGs to estimate surprisal, which could presumably make substantially better predictions for words via interpolation with $n$-gram models.

Experiment 2 investigated whether $n$-gram probability had significant effects on gaze duration, beyond those necessary to improve perplexity of a language model. For current word processing, the results showed significant effects of unigram, bigram, and trigram probability above and beyond surprisal, to a much greater extent than in Experiment 1. This was seen in much larger interpolation weights for \textit{n}-grams, and the resultant perplexities from these weightings, which were far from optimal for language model-based perplexity. On the other hand, for the prior word, only word frequency (unigrams) showed an effect much beyond surprisal.

These results provide strong evidence that word frequency is not the only local language statistic to affect processing above and beyond surprisal, but rather, we also see effects of two other local statistics, bigrams and trigrams. 

One possible objection to these conclusions is that the language models used in this study are not representative of the actual (implicit) language models that humans are effectively using to predict and comprehend text, and thus we cannot draw strong claims about human language processing from them. While we agree that these computational language models are clearly not the same as those used by humans (since, e.g., humans are still much better at making contextualized predictions for upcoming words than any computational language model), we argue this is not a strong limitation on our conclusions. First, as demonstrated in Expt.~1, the fully contextualized language model we used does a reasonably good job of predicting the words in the test set (perplexities below 100), which makes sense given that both training and test sets are composed of text from news media. Second, it has been established that (within common classes of language models) there is a relatively tight relationship between the \emph{linguistic accuracy} of a language model, measured by perplexity, and its \emph{psychological accuracy} or predictive power for human reading times  \citep{monsalve2012lexical,goodkind2018predictive}, with some of the evidence coming from this very corpus. Given these findings, we know of no reason to believe that interpolating a high quality language model with local statistics, such that they hurt the model's predictive power, would make its predictions more correlated to human reading times.

Rather, we argue that these findings represent strong evidence that natural reading times exhibit outsized effects of local statistics. The question remains, however, \emph{why} this is the case. These results can be interpreted in two broad ways. One class of possibilities is that comprehenders make use of systematically \emph{irrational} processing strategies that rely more on local information than would be ideal \citep[e.g.,][]{tabor2004effects}, due, for example, to some biases in the language processing system.
An alternative class of interpretations is that outsized effects of local statistics arise because comprehenders are rationally utilizing imperfect, ``noisy'' representations of linguistic context \citep{levy2008noisy}.

One instance of the irrational hypothesis that might be able to explain our results could be that there are effects of local perceptual learning on human sentence processing. For example, perhaps the visual system has adapted over experience to be faster at recognizing words in particular visual contexts, e.g., in frequent bigram phrases. If such a hypothesis were true, we would expect independent effects of bigram probability (arising from the visual system biases) and effects of surprisal given the full language context (arising from the language system), which could thus explain our results. Of course, many other instances of the irrational possibility also exist, such as the language system itself being irrationally sensitive to local context \citep[e.g.,][]{tabor2004effects}, or frequent word sequences being stored or accessed in some privileged way \citep[e.g.,][]{arnon2010more} that goes beyond what would be optimal.

On the other hand, an instance of the rational hypothesis that might be able to explain our results comes from noisy-context or ``lossy'' surprisal \citep{futrell2017noisy,futrell2020lossy}, an extension of classic surprisal. In noisy-context surprisal, information that comprehenders have about the linguistic context is corrupted by noise, such as imperfect memory representations. Thus, while all previous context is in principle still taken into account during processing, the effect of a particular piece of context will depend on how much noise has been applied to it. In simple variants of this model, processing each new word will add some noise to preceding words, and so a natural consequence is that the local context will have the least noise applied, and thus outsized effects on processing. 

This type of theory would thus also predict our finding of simultaneous effects of $n$-gram probability and surprisal on reading times. In contrast to the irrational hypothesis mentioned previously, however, this theory would predict these effects not from two independent sources (vision and language) but from one complex process, in which local information is just overweighted due to noise.

To test whether a theory such as noisy-context surprisal, perceptual learning, or some other theory, is responsible for the observed results requires further study. One way of contrasting these theories would be to examine differences between effects of $n$-gram models of different sizes, or to look at interactions with word length. Because visual quality decreases quite rapidly away from fixation, a perceptual learning account should not predict effects of words far from fixation.

\section{Conclusion}
Whatever the explanation, in this paper we have presented evidence that local statistics affect reading times above and beyond surprisal given the full context. These are not results that the classic surprisal theory can explain, with its causal bottleneck. Instead, our findings suggest the need for exploring new unifying computational theories of human sentence processing.

\section*{Acknowledgements}
This work was supported by NSF grant 1734217.


\bibliography{acl2020.bib}
\bibliographystyle{acl_natbib}

\end{document}